%% file: paper.tex
\documentclass{article} 
\usepackage{nips13submit_e,times}
\usepackage{hyperref}
\usepackage{url}
\usepackage{amsmath}
\usepackage{graphicx}

\include{macros}

\title{Inferring Multilateral Relations from\\Dynamic Pairwise Interactions}

\author{
Aaron Schein\thanks{Aaron Schein and Juston Moore are joint principal authors.}, Juston Moore, Hanna Wallach\\
School of Computer Science\\
University of Massachusetts Amherst\\
Amherst, MA 01003\\
\texttt{\{aschein, jmoore, wallach\}@cs.umass.edu}
}

\nipsfinalcopy 

\begin{document}

\maketitle

\begin{abstract}
Correlations between anomalous activity patterns can yield pertinent information about complex social processes: a significant deviation from normal behavior, exhibited simultaneously by multiple pairs of actors, provides evidence for some underlying relationship involving those pairs---i.e., a \emph{multilateral relation}. We introduce a new nonparametric Bayesian latent variable model that explicitly captures correlations between anomalous interaction counts and uses these shared deviations from normal activity patterns to identify and characterize multilateral relations. We showcase our model's capabilities using the newly curated Global Database of Events, Location, and Tone, a dataset that has seen considerable interest in the social sciences and the popular press, but which has is largely unexplored by the machine learning community. We provide a detailed analysis of the latent structure inferred by our model and show that the multilateral relations correspond to major international events and long-term international relationships. These findings lead us to recommend our model for any data-driven analysis of interaction networks where dynamic interactions over the edges provide evidence for latent social structure.
\end{abstract}

\section{Introduction}

Social processes are complex because relationships between actors are not independent. In many domains, we only observe interactions between pairs of actors, but we are interested in learning about the latent structure underlying these pairwise relationships. For example, in international relations, treaties and coalitions influence the pairwise behavior of countries. A particular type of multilateral relation, a coalition, can result in a group of countries taking action toward another country. We are interested in characterizing \emph{multilateral relations}, persistent relationships between pairs of actors in an interaction network.
When multilateral relations are inherently latent, temporal correlations in interactions across edges in the network provide evidence for the underlying relations.





%

%

Every pair of actors has its own inherent normal rate of activity, which is not influenced by multilateral relations. Machine learning researchers have traditionally relied on normal activity patterns in order to infer latent structural information. There are many applications in which correlations in anomalous behavior yield more pertinent information than correlations in normal activity patterns. 
A significant deviation from normal behavior, exhibited by multiple pairs of actors, provides evidence for an underlying multilateral relation. Pairs involved in the same multilateral relations are likely to show similar anomalous deviations from their normal interaction rates at the same points in time.
Although there are existing models for anomaly detection in complex social processes, most do not model similarities between actors and cannot detect multilateral relations. For example, Heard et al.’s model for anomaly detection in social networks \cite{heard2010bayesian} assumes that the events associated with each actor are independent, and therefore ignores any correlations between actors' anomalous activity patterns. Similarly, Kleinberg’s model for identifying periods of bursty observation counts \cite{kleinberg2003bursty} finds only those events involving a single actor. We present a new nonparametric Bayesian latent variable model that explicitly captures correlations between anomalous interactions of pairs, thereby facilitating the detection and characterization of multilateral relations.


We model dynamic interactions between actors as a sequence of weighted graphs, where actors are represented as nodes and the weights on each directed edge represent the number of directed interactions in a particular time slice. This representation is equivalent to considering a single graph with edges annotated with a time series of interaction counts over discrete time slices. Grouping pairs based on anomalous interactions is equivalent to clustering annotated edges in this framework.  A large amount of work in network science has been devoted to discovering groups of nodes in a network based on structural properties of the network, where edges are either binary or weighted.  These clustering algorithms assign a node to a single cluster \cite{newman2006modularity,kemp2006learning} or infer a mixed-membership cluster assignment \cite{airoldi2008mixed}. The existing work largely ignores attributes that might be attached to edges, and in particular doesn't include dynamics of interactions over edges. Instead, we choose to directly model the dynamics of interactions over edges. Our model explicitly groups the edges in an interaction network based on the dynamics of the associated time series. The resulting groups can be interpreted as mixed-membership assignments for the nodes by associating a node with every relation to which its outgoing edges are assigned.  However, considering groupings of edges allows us to discover more nuanced patterns of interaction.





We describe the mathematical details of our new model in the next section and outline an efficient procedure for approximate inference in section~\ref{sec:Inference}. In section \ref{sec:Synthetic} we validate our inference procedure by demonstrating that it can recover the parameters used to generate synthetic data.  In section~\ref{sec:Exploratory}, we apply our model to international relations data consisting of directed pairwise interactions automatically extracted from news articles from around the world between 1979 and the present, newly curated in the Global Database of Events, Language, and Tone (GDELT) \cite{kaley2003gdelt}.  This data is of interest to researchers in the social sciences, as well as government and business intelligence analysts because it provides a high-level view of the relationships between countries over the last several decades.  We first show that our model discovers important international events that occurred during the last decade and further provide a detailed analysis of the latent structure inferred by the model.  




\section{Model}

In this section, we present our new nonparametric Bayesian latent variable model for discovering multilateral relations from periods of correlated anomalous pairwise interaction.


We model pairwise interactions occurring over directed edges between a set of $N$ actors. Each directed edge from actor $i$ to actor $j$ is associated with a $T$-length vector of interaction counts, $\boldsymbol{y}_{ij} = \left\lbrace y_{ij}^{(t)} \right\rbrace_{t=1}^T$, recording the number of times $i$ acted toward $j$ in each discrete time slice $t$. We assume that edges are partitioned into latent groups. Each edge $(ij)$ has a group assignment $z_{ij}$; two edges are in the same group if their group assignments are the same. Given its group assignment, each edge draws a number of interactions counts at each time slice $t$, from a Poisson distribution parameterized by the product of an edge-specific interaction base rate $\lambda_{ij}$ and a group- and time-specific deviation factor $\delta_{z_{ij}}^{(t)}$: that is, $y_{ij}^{(t)} \sim $ Pois$\left( \lambda_{ij} \, \delta_{z_{ij}}^{(t)} \right)$. The base rate for each edge describes the expected number of interactions inherent to that edge during periods of normal activity.  Each base rate is drawn from a global Gamma prior, with shape and scale parameters $\boldsymbol{\gamma} = \left\lbrace \gamma_1, \gamma_2 \right\rbrace$; we in turn assume a noninformative prior over each of these hyperparameters.  When a group's deviation factor $\delta_{z_{ij}}^{(t)}$ is equal to one, the count of interactions occurring over each edge $(ij)$ in that group is drawn from a Poisson distribution with mean equal to the base rate, $\lambda_{ij}$; when the deviation factor increases (or decreases) from one, the mean of the Poisson distribution is increased (or decreased) accordingly. The deviation factors for each group therefore capture shared temporal correlations in the counts of interactions occurring over the edges in that group.

We assume that most of the deviation factors are close to one, and that only the anomalous time slices correspond to deviation factors much greater than or less than one. Our model considers each deviation factor $\delta_{g}^{(t)}$ to be a drawn from one of two types of distributions: a \emph{spike} distribution, which is highly peaked around one, and one or more \emph{slab} distributions, which have support above or below one in order to model anomalously high counts, or low counts, respectively. In order to capture periods of anomalous activity, we encode the assumption that anomalies follow anomalous time slices, and normal time slices follow other normal time slices by modeling the deviation factors $\boldsymbol{\delta}_g = \left\lbrace \delta_g^{(t)} \right\rbrace_{t=1}^T$ as ``observations'' from a $K$-state hidden Markov model (HMM). The latent states are denoted by $\boldsymbol{s}_g = \left\lbrace s_g^{(t)} \right\rbrace_{t=1}^T$, and the transition probabilities are encoded in a matrix $\Theta$, where the entry $\theta_{ss'}$ gives the probability of a transition from state $s$ at time $t$ to state $s'$ at time $t+1$. We place an asymmetric Dirichlet prior distribution over each row in $\Theta$ (parameterized by a vector $\boldsymbol{\beta}_s$), and assume that the first state $s_g^{(1)}$ is preceded by $s_g^{(0)} = 0$. The HMM emission distribution for state $s_g^{(t)}=0$ is a low-variance spike distribution so that all corresponding deviation factors will be close to one. The emission distributions corresponding to the other states $s_g^{(t)} \in \left\lbrace 1, \ldots, K-1 \right\rbrace$ provide support greater than, or less than, one to model increased or decreased values of $\delta_g^{(t)}$, respectively. Each emission distribution for state $s$ is assumed to be a Gamma distribution with shape parameter $c_s$ and scale parameters $d_s$.

Since it is often impossible to determine the number of multilateral relations underlying observed interactions, we allow the model to infer an appropriate number of groups. The partition of edges implied by their group assignments $\mathcal{Z} = \{ z_n \}_{n=1}^N$ is drawn from a Chinese restaurant process prior~\cite{pitman2002combinatorial} with concentration parameter $\alpha$. By determining the number of groups in the partition, this prior causes edges to be grouped together only when they exhibit anomalous deviations from their normal interaction base rates at similar points in time.

\section{Inference}
\label{sec:Inference}

Given an observed time series of interaction counts for each edge in an interaction network, $\mathcal{Y} = \left\lbrace \boldsymbol{y}_{ij} \right\rbrace_{i \ne j}$, and hyperparameters $\mathcal{H} = \left\lbrace \left\lbrace \boldsymbol{\beta}_s, c_s, d_s \right\rbrace_{s=1}^K, \alpha \right\rbrace$, our model permits inference of the edge-specific interaction base rates $\Lambda = \left\lbrace \lambda_{ij} \right\rbrace_{i \ne j}$, the group-and time-specific deviation factors $\Delta = \left\lbrace \boldsymbol{\delta}_g \right\rbrace_{g=1}^G$, the states of the Hidden Markov Model $\mathcal{S} = \left\lbrace \boldsymbol{s}_g \right\rbrace_{g=1}^G$, and the group assignments $\mathcal{Z} = \left\lbrace z_{ij} \right\rbrace_{i \ne j}$. For convenience, we will index the groups in the partition implied by $\mathcal{Z}$ as $g \in \left\lbrace 1, \ldots, G \right\rbrace$.

Gamma-Poisson conjugacy allows us to marginalize out all the deviation factors, giving the following expression for the joint distribution over the observed counts and the latent variables of interest\footnote{The P\'{o}lya distribution is also known as the Dirichlet-multinomial distribution.}:
\begin{align}
&P\left(\mathcal{Y}, \Lambda, \mathcal{S}, \mathcal{Z} \mid \mathcal{H}\right) = \int \text{d}^{G \times T} \Delta \; P\left(\mathcal{Y}, \Lambda, \Delta, \mathcal{S}, \mathcal{Z} \mid \mathcal{H}\right) \notag\\
	&\quad= \prod_{g=1}^G\prod_{t=1}^T \left( \frac{\left(d_{s_g^{(t)}}\right)^{-c_{s_g^{(t)}}}}{\Gamma\left(c_{s_g^{(t)}}\right)} \frac{\Gamma\left({c'}_g^{(t)}\right)}{\left({d'}_g^{(t)}\right)^{-{c'}_g^{(t)}}} \prod_{(ij) \in g} \left( \frac{\lambda_{ij}^{y_{ij}^{(t)}}}{\Gamma\left(y_{ij}^{(t)}+1\right)} \right) \right) \notag\\
	&\quad\quad \prod_{n=1}^N \text{Gamma}\left(\lambda_n \g \boldsymbol{\gamma} \right) \prod_{s=1}^S \text{P\'{o}lya}\left(\left\langle n_{s1}, \ldots, n_{sK} \right\rangle \g \boldsymbol{\beta}_{s} \right) \text{CRP}(\mathcal{Z} \g \alpha)
\end{align}
where ${c'}_g^{(t)} = c_{s_g^{(t)}} + \sum_{(ij) \in g} y_{ij}^{(t)}$ and ${d'}_g^{(t)} = \left(\frac{1}{d_{s_g^{(t)}}} + \sum_{(ij) \in g} \lambda_{ij}\right)^{-1}$ are sufficient statistics, and $n_{ss'} = \sum_{g=1}^G \sum_{t=2}^T \boldsymbol{1}\left(s_g^{(t-1)} = s\right)\,\boldsymbol{1}\left(s_g^{(t)} = s'\right)$ is the total number of state transitions from state $s$ to state $s'$ in the HMM.\footnote{The function $\mathbf{1}(\cdot)$ evaluates to one if its argument evaluates to true and evaluates to zero otherwise.}
The term CRP refers to the likelihood of a partition drawn from a Chinese Restaurant Process, given by:
\begin{align}
\text{CRP}(\mathcal{Z} \g \alpha)
	&= \frac{\Gamma(\alpha)}{\Gamma(\sum_{g=1}^G n_g + \alpha)} \alpha^{G} \prod_{g=1}^G \Gamma(n_g),
\end{align}
where $n_g$ is the number of entities associated with group $g$ and $G$ is the number of groups in the partition implied by $\mathcal{Z}$.

We use Markov chain Monte Carlo to sample typical values for each of latent variables in our model from their joint posterior given $\mathcal{Y}$. In each iteration, we sample a value for each base rate $\lambda_{ij}$ as well as typical hyperparameter values for $\gamma_1, \gamma_2$ from their posterior distributions using a slice sampling algorithm~\cite{neal2003slice}. We sample each group assignment $z_n$ from its conditional posterior using Neal et al.'s ``Algorithm 8'' for Gibbs sampling from a nonconjugate Dirichlet process~\cite{neal2000markov}. We draw a sample for the HMM transition matrix $\Theta$ from its posterior distribution conditioned on $\mathcal{S}$, using Dirichlet-multinomial conjugacy. Finally, we jointly sample values for $\boldsymbol{s}_g = \left\lbrace s_g^{(1)}, \ldots, s_g^{(T)} \right\rbrace$ from their exact conditional distribution by applying the forward-filtering backward-sampling algorithm for inference in HMMs~\cite{murphy2012machine}.




\section{Results}

In this section we describe synthetic experiments that establish the validity of our model and its approximate inference algorithm, and we showcase our model's exploratory capabilities by performing an in-depth analysis of a large interaction network from international relations.

\subsection{Synthetic experiments}
\label{sec:Synthetic}

We generated four synthetic datasets, with the same setting of parameters\footnote{We set the CRP concentration parameter $\alpha$ to 1.0.  We set the shape $\gamma_{1}$ and scale $\gamma_{2}$ hyperparameters for the Gamma prior over edge-specific base rates to 2 and 5, respectively.  We specified a two-state HMM over activation states and set the Dirichlet prior over the transition matrix to $\rho_{00}=800, \rho_{01}=80, \rho_{10}=200, \rho_{11}=600$.  For the case of $s_{g}^{(t)} = 0$, we set the emission distribution over $\delta_{g}^{(t)}$ to a Gamma distribution with shape $c_{0}$ and scale $d_{0}$ equal to 1000 and 0.0001 (the \emph{spike}).  For the case of $s_{g}^{(t)}=1$, we set the shape $c_{1}$ and scale $d_{1}$ to 4 and 1 (the \emph{slab}).}, for all combinations of 1000 or 2000 edges and 52 or 365 time slices (corresponding to daily and weekly binning of a single year).  We selected numbers of actors and time slices, as well as hyperparameters, that would generate synthetic data similar to the real data we are interested in modeling.  In each synthetic experiment, we ran our model over one of these four synthetic datasets and plotted the error in the inferred latent variables under various settings of the hyperparameters. The results show that our model is capable of recovering the latent parameters in these synthetic datasets. In each synthetic experiment, we ran our model over one of these four synthetic datasets and plotted the error in the inferred latent variables under various settings of the hyperparameters to confirm that we recovered the true parameterization.  We display an example of these plots in Figure 1.

\begin{figure}
\centering
\includegraphics[width=0.8\textwidth]{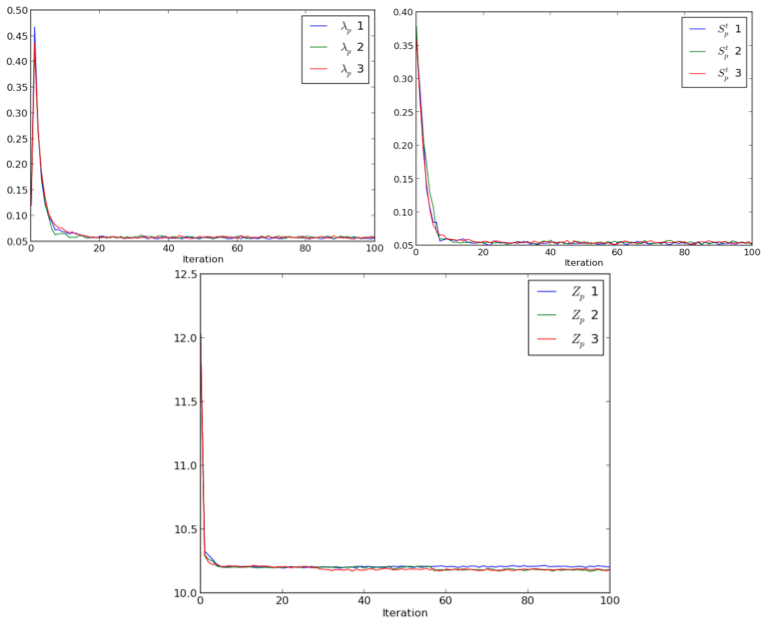}
\caption{Three plots showing error over sampling iterations.  Top left is error on base rates, top right is error on activation states, bottom is variance of information on group assignments.  Inference in this experiment was performed conditioned on the true values for all hyper parameters expect for the CRP concentration parameter which was different (0.01 instead of 120).}
\end{figure}

\subsection{Global Database of Events, Language, and Tone (GDELT)}

GDELT \cite{kaley2003gdelt} is a new data set consisting of over 200 million records of real-world events from 1979 to present.  Each record contains the basic core elements that identify a unique event -- who did what do whom, when, and where -- that are extracted automatically from news articles.  From this event record data, we induce an interaction network between country actors by counting the total number of articles involving directed actions from one country to another in each discrete time slices.
We expect that anomalously high counts of interaction for some pair of countries corresponds to some event of real significance involving that relation, since an anomalously high count means that the media has suddenly redirected its focus.  By correlating these increases in media attention, our model finds edges exhibiting anomalously high interaction counts at similar times due to underlying multilateral relations.

In international relations, multilateral relations change (sometimes drastically) over time -- however, we assume they are constant within a short enough time window.  In our exploration of GDELT, we ran our model to infer a partition of the most active 1,000 edges independently for each of the years 2000 to 2012. The model was run with two HMM states describing deviation factors, a spike distribution with mean 1 and variance 0.001, and a slab distribution with mean 4 and variance 16. Each run consisted of 3 independent MCMC chains, each with 5,000 sampling iterations.  For each year, we identified the single sample across all 3 chains with the highest joint probability and inspected the groups in its inferred partition in decreasing order of size.  The groups we found tended either to correspond to multilateral relations that formed temporarily around a major event or ones that are longterm and based around some geopolitical context shared between its edges.  Some of the groups corresponded to well-known events about which the authors had prior knowledge.  Others, however, required us to learn about some interesting and significant real-world events.  We consider this an important validation of the model as an exploratory tool and describe some of these exploratory results in the following section.

\subsection{Exploratory Results}
\label{sec:Exploratory}

\begin{figure}
\label{fig:a}
\vspace{-4mm}
\caption{Central Asian republics: This group contains all of the directed edges between the former Soviet-bloc Central Asian republics: Kyrgyzstan, Tajikistan, Uzbekistan, Kazakhstan.  Our model inferred a very similar group across all the years.  Indeed, this is an example of a multilateral relationship that exists due to a longterm shared geopolitical context.}
\includegraphics[width=\textwidth]{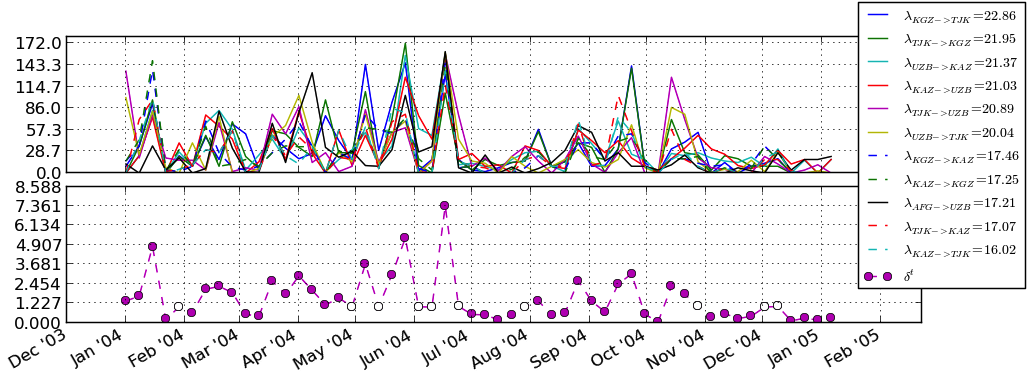}
\caption{2004 enlargement of the European Union (as it is called on Wikipedia):  Several European countries simultaneously acceded to the EU in May 2004 including Poland, Slovakia, Latvia, and Estonia.  This group contains directed edges between those countries and EU, many of which exhibit a burst in activity during and surrounding that time.}
\includegraphics[width=\textwidth]{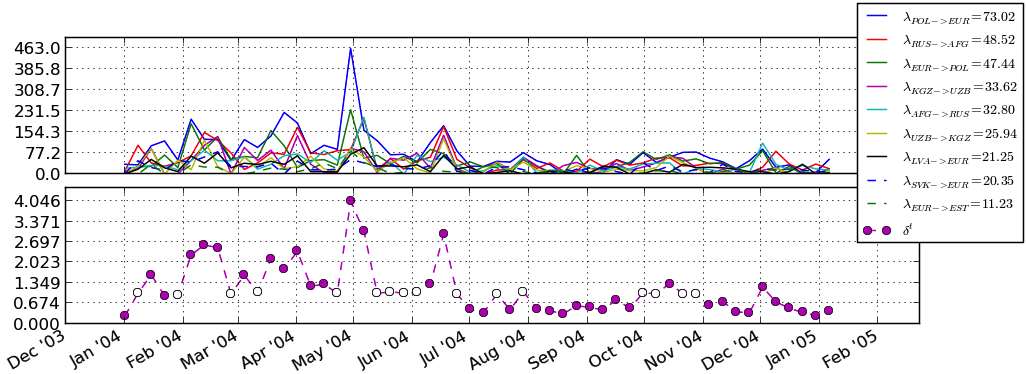}
\caption{2004 Indian Ocean tsunami: A single event in December clearly ties these edges together -- this is the tsunami that hit Southeast Asia in 2004 and resulted in a massive humanitarian crisis and international response.  The edges involve Pacific countries close to the disaster: China, Japan, Australia, Philippines.}
\includegraphics[width=\textwidth]{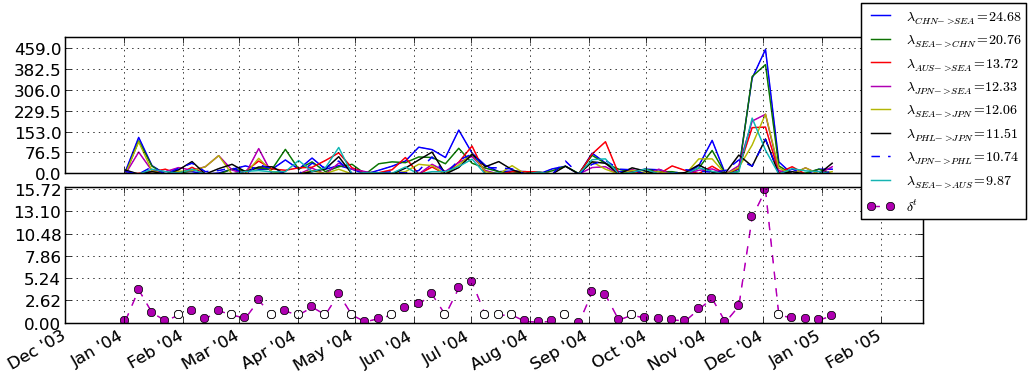}
\end{figure}

\begin{figure}
\label{fig:b}
\vspace{-4mm}
\caption{Tulip Revolution: In April 2005, a popular uprising in Kyrgyzstan, later called the ``Tulip Revolution'', overthrew the government of President Askar Akayev.  This group contains edges involving Kyrgyzstan, the most active ones being those with the Russia and the United States. While Russia has always maintained an interest in the neighboring Central Asian republics, the United States had a particular stake in maintaining its access to a central military installation near the capital, Bishkek, that was critical for servicing its ongoing operations in Afghanistan.  }
\includegraphics[width=\textwidth]{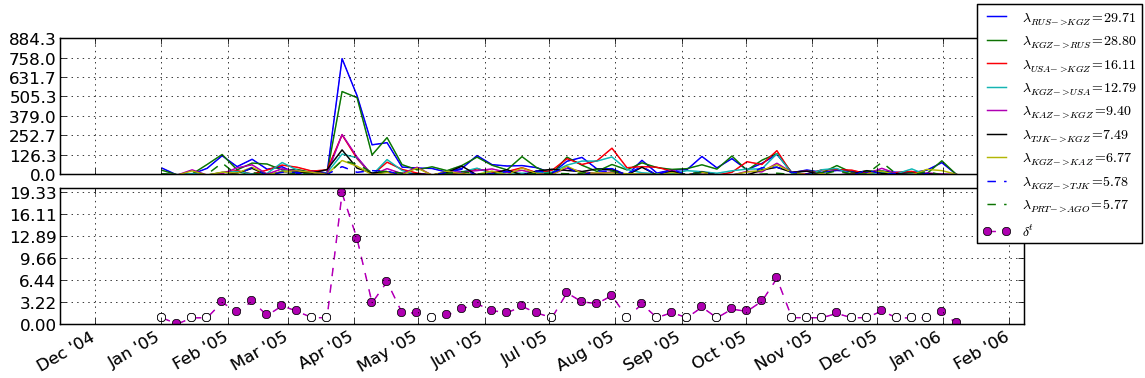}
\caption{2005 terrorist attacks in London: In July 2005, suicide bombers attacked London's public transportation system.  Though the terrorists were home-grown, the attacks garnered an international response particularly from Coalition countries in the War on Terror.  Most of the edges in this group involve Great Britain and exhibit a particular spike of activity during the time of the bombings. }
\includegraphics[width=\textwidth]{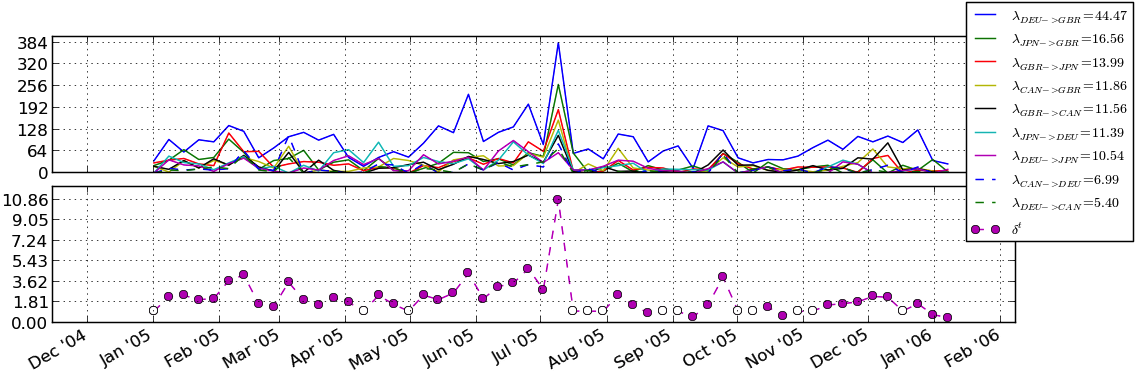}
\caption{Negotiations with Iran: Throughout 2005, Iran was involved in negotiations with several European countries (mainly France and Germany) over its controversial nuclear program.  This is a classic example of a multilateral relation in international relations: a multiparty negotiation. }
\includegraphics[width=\textwidth]{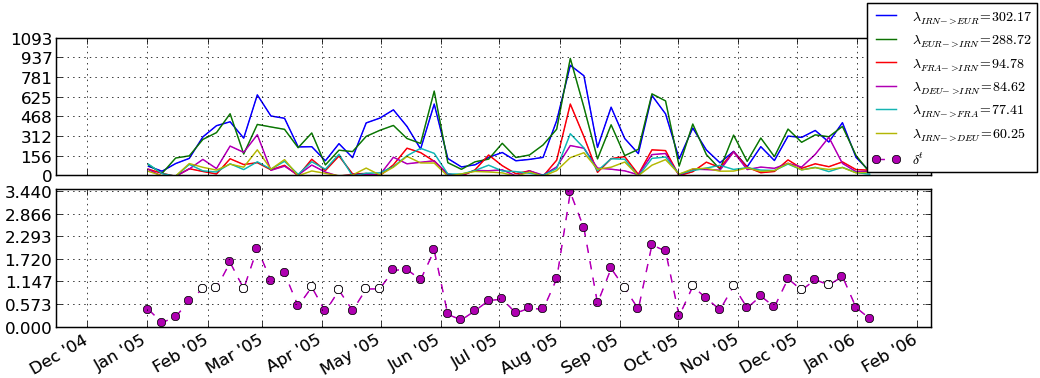}
\end{figure}

For an exploratory tool to be useful, it must be capable of discovering significant and previously unknown patterns in data.  For such a tool to be trusted by practitioners though, it must also discover ``obvious'' patterns, that are consistent with prior knowledge \cite{sim2013measuring}.  We formulated mini-hypotheses about certain ``obvious'' groups we expected to see based on our knowledge of major international events, and we confirmed that our model inferred groups corresponding to those events.  For example, after running inference on the 2003 data, we observed that several of the largest groups inferred by the model corresponded to relations involved in the invasion of Iraq (confirming our mini-hypothesis that this major event should be represented).  

We show 3 of the largest 5 groups for each of the 2004 and 2005 runs in Figures 2-4 and 5-7.  Some of these groups correspond to multilateral relations that formed temporarily around major international events while others correspond to longterm multilateral relations formed around shared geopolitical contexts.  One of the groups we display -- the Central Asian republics -- is an example of a longterm multilateral relation: though we only display the group inferred from the 2004 run, our model inferred a very similar group across all of the years. We didn't know about some of the groups -- this is what an exploratory tool is for.  In each the figures on the following two pages, we plot the time series for all of the edges in a single group in the top subplot.  In the bottom subplot, we plot the posterior mean of that group's deviation factor $\delta_{g}^{(t)}$, at each time slice. The deviation factors at each time slice are colored white or red, denoting the state of the HMM at for the corresponding group at that time slice, corresponding to the spike distribution ($s_{g}^{(t)} = 0$) and the slab distribution ($s_{g}^{(t)} = 1$), respectively.




\subsubsection*{Acknowledgments}

We would like to thank University of Massachusetts Amherst professors Benjamin Marlin and Daniel Sheldon for many helpful conversations, and feedback on our work.

\section{Conclusions}

We introduced a new non-parametric Bayesian latent variable model that discovers latent structure in complex social processes by capturing correlated deviations from normal behavior.  We demonstrated that our model recovers the true parameters of simulated data and presented the results from running our model on a subset of the Global Database of Events, Language, and Tone, an exciting data set that is new to the machine learning community.  Our findings from the country interaction data are consistent with established theory in international relations.  Based only on the observed interactions between pairs of countries, our model discovers underlying multilateral relations that explain correlated deviations from normal behavior and are consistent with our prior knowledge of international affairs.

\bibliographystyle{unsrt}

\small{
\bibliography{nips13.bib}
}

\end{document}

%% file: macros.tex
\newcommand{\comment}[1]{}

\newcommand{\g}{\,|\,}
